\documentclass{article}

\usepackage{microtype}
\usepackage{graphicx}
\usepackage{varwidth}
\usepackage{subcaption}
\usepackage{booktabs} % for professional tables
\usepackage{hyperref}

\usepackage[preprint]{icml2026}
\usepackage{tabularx}
\usepackage{amsmath}
\usepackage{amssymb}
\usepackage{mathtools}
\usepackage{amsthm}
\usepackage{multirow}
\usepackage{subcaption}

\usepackage[capitalize,noabbrev]{cleveref}

\theoremstyle{plain}

\theoremstyle{definition}

\theoremstyle{remark}

\DeclareMathOperator*{\argmax}{arg\,max}

\usepackage[textsize=tiny]{todonotes}

\icmltitlerunning{On Representation Redundancy in Large-Scale Instruction Tuning Data Selection}

\begin{document}

\twocolumn[
  \icmltitle{On Representation Redundancy in \\ Large-Scale Instruction Tuning Data Selection}



  \icmlsetsymbol{equal}{\textdagger}

  \begin{icmlauthorlist}
    \icmlauthor{Youwei Shu}{nus,equal}
    \icmlauthor{Shaomian Zheng}{ant}
    \icmlauthor{Dingnan Jin}{ant}
    \icmlauthor{Wenjie Qu}{nus}
    \icmlauthor{Ziyao Guo}{nus,equal} \\
    \icmlauthor{Qing Cui}{ant}
    \icmlauthor{Jun Zhou}{ant}
    \icmlauthor{Jiaheng Zhang}{nus}
  \end{icmlauthorlist}

  \icmlaffiliation{nus}{National University of Singapore}
  \icmlaffiliation{ant}{Ant Group}

  \icmlkeywords{Machine Learning, ICML}

  \vskip 0.3in
]



\printAffiliationsAndNotice{\icmlEqualContribution}

\begin{abstract}
  Data quality is a crucial factor in large language models training. While prior work has shown that models trained on smaller, high-quality datasets can outperform those trained on much larger but noisy or low-quality corpora, systematic methods for industrial-scale data selection in instruction tuning remain underexplored. In this work, we study instruction-tuning data selection through the lens of semantic representation similarity and identify a key limitation of state-of-the-art LLM encoders: they produce highly redundant semantic embeddings. To mitigate this redundancy, we propose Compressed Representation Data Selection (CRDS), a novel framework with two variants. CRDS-R applies Rademacher random projection followed by concatenation of transformer hidden-layer representations, while CRDS-W employs whitening-based dimensionality reduction to improve representational quality. Experimental results demonstrate that both variants substantially enhance data quality and consistently outperform state-of-the-art representation-based selection methods. Notably, CRDS-W achieves strong performance using only 3.5\% of the data, surpassing the full-data baseline by an average of 0.71\% across four datasets. Our code is available at \url{https://github.com/tdano1/CRDS}.
\end{abstract}

\section{Introduction}

Instruction tuning, also known as supervised fine-tuning (SFT) \cite{DBLP:conf/nips/Ouyang0JAWMZASR22}, is a crucial post-training stage for large language models (LLMs), as it endows base models with instruction-following and dialogue capabilities. Unlike the vast and heterogeneous corpora used during pre-training, SFT relies on a substantially more focused dataset. Nevertheless, many recent LLMs \cite{DBLP:journals/corr/abs-2407-21783,DBLP:journals/corr/abs-2412-15115,DBLP:journals/corr/abs-2510-22115} still require millions of SFT examples to effectively activate the capabilities of the underlying base model, making data curation both critical and challenging. On the one hand, equipping the model with strong conversational abilities while ensuring robustness against safety risks \cite{DBLP:journals/corr/abs-2212-08073, DBLP:journals/corr/abs-2307-09288, DBLP:journals/tmlr/YangHBLQZBWYSTSRLWZCWZZ25, DBLP:conf/uss/0001ZTYJTZ0Z25} demands that the SFT dataset be both comprehensive and of high quality. On the other hand, excessive data volume can introduce conflicting or inconsistent knowledge, potentially degrading model performance across multiple dimensions. From a broader perspective, beyond establishing basic inference and conversational abilities, instruction tuning serves as a critical intermediate step in modern LLM training pipelines, significantly influencing the effectiveness of subsequent post-training stages such as reinforcement learning \cite{DBLP:journals/corr/abs-2203-02155} and direct preference optimization \cite{DBLP:conf/nips/RafailovSMMEF23}. 

  \begin{figure*}[t]
  \centering
  \includegraphics[width=\textwidth]{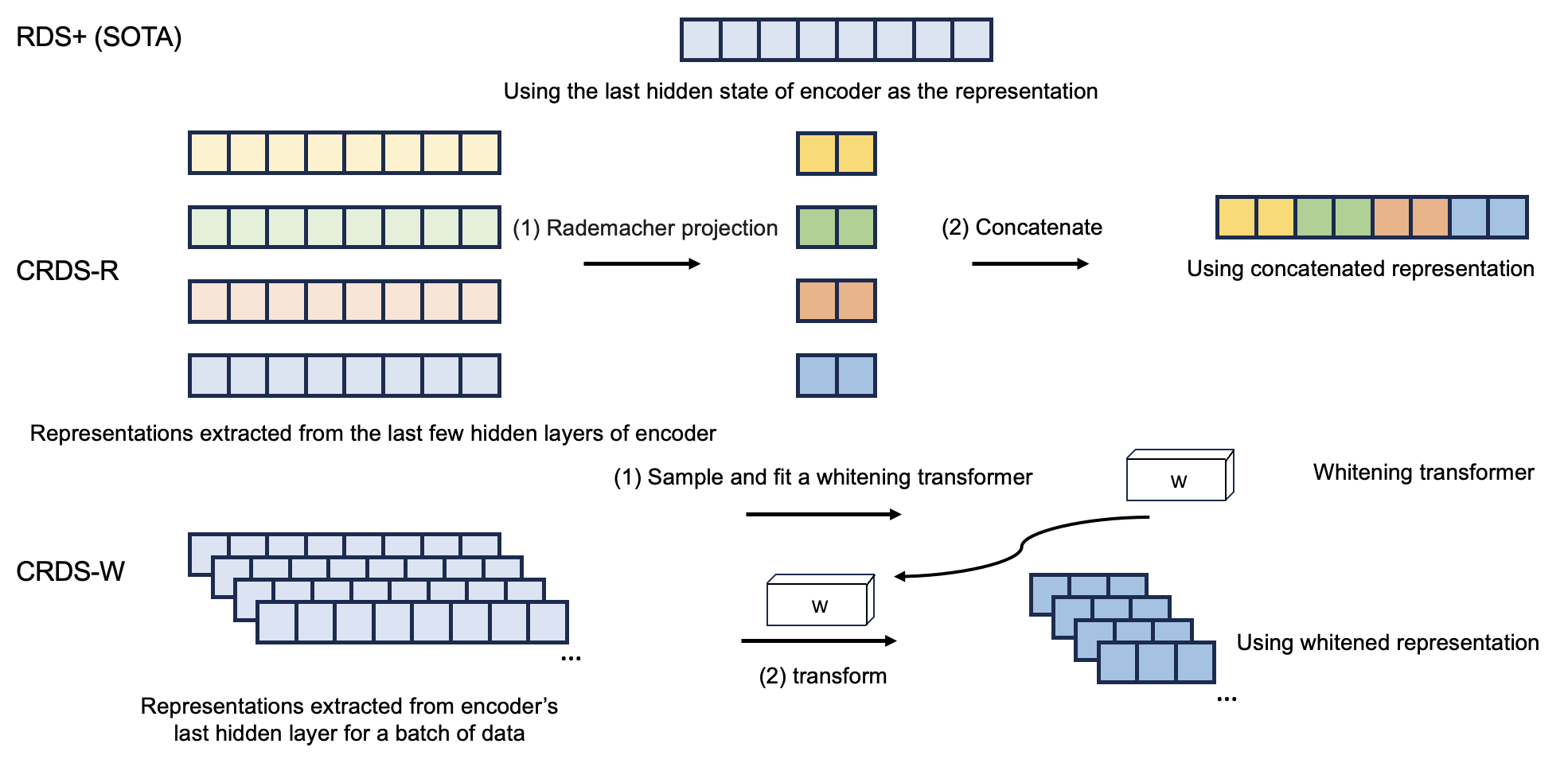}
  \caption{Illustration of our proposed methods and the SOTA baseline RDS+.
RDS+ constructs data representations by directly using the last hidden state of the encoder.
CRDS-R constructs representations by extracting the last several hidden states of the encoder, applying a Rademacher projection to each layer, and concatenating the projected features to obtain the final representation, which retains the same dimensionality as the original representation.
CRDS-W constructs representations by first fitting a whitening transformer on a large subset of the data using the last hidden state of the encoder, and then applying this transformer to whiten all data representations for similarity computation.}
  \label{fig:pipeline}
  \vspace{-0.3cm}
\end{figure*}

Even before the advent of LLMs, data selection, also called data pruning and data condensation, had already attracted significant attention from researchers \cite{DBLP:conf/icml/MirzasoleimanBL20,DBLP:conf/nips/SorscherGSGM22,DBLP:conf/iclr/ZhaoMB21}, as it became widely recognized that data quality profoundly impacts model training. With the rise of LLMs, numerous studies have demonstrated that the scale of training datasets can often be drastically reduced without compromising performance \cite{DBLP:conf/nips/ZhouLX0SMMEYYZG23, DBLP:journals/corr/abs-2502-03387}. Recently, a substantial body of work has focused on instruction tuning data selection, particularly on standard small datasets such BBH \cite{DBLP:conf/acl/SuzgunSSGTCCLCZ23} as GSM8K \cite{DBLP:journals/corr/abs-2110-14168}. However, in industrial settings, datasets often span millions of examples or more—a scale at which many methods that excel on smaller datasets tend to lose their effectiveness \cite{DBLP:journals/corr/abs-2503-01807}.

Currently, a notable gap remains in the literature: very few approaches are specifically designed to address data selection challenges at an industrial scale. Most existing methods focus on selecting data from a single instruction-tuning dataset, which typically contains on the order of $10^4$ to $10^5$ examples. Furthermore, the majority of current techniques rely on human-designed intuitive metrics \cite{DBLP:conf/iclr/LuY0LLTZZ24, DBLP:conf/iclr/ChenLYWGYTS0HJ24, DBLP:conf/iclr/0131Z00H24, DBLP:conf/nips/Liu0W0WH024, DBLP:conf/nips/LiuKR24, DBLP:conf/nips/HeWSSDW0024}. Even when similarity-based algorithms are considered, most methods adopt gradient-based criteria as their primary signal \cite{DBLP:conf/icml/XiaMGA024, DBLP:conf/acl/LinLX024}, which are computationally expensive. Although many of these approaches achieve strong performance—sometimes even outperforming full-data baselines at significantly reduced scales—they remain difficult to scale. In contrast, semantic representations, as a primitive yet convenient property of data, have not been sufficiently explored as a criterion for instruction-tuning data selection, and their effectiveness has been largely underestimated.

To address this gap, we systematically investigate the effectiveness of semantic representations for instruction-tuning data selection. We find that while semantic embeddings are highly informative, they also exhibit substantial redundancy. We further demonstrate that two dimensionality reduction techniques—Rademacher projection and whitening—are remarkably effective for instruction-tuning data selection. This finding is significant for two reasons. First, we demonstrate that a Rademacher projection–concatenation pattern applied to transformer hidden representations yields a superior feature space for data selection. Second, we demonstrate that classical dimensionality reduction methods can be applied to instruction-tuning data selection in a straightforward yet effective manner. Together, these results substantially amplify the practical advantages of instruction tuning. Compared to pretraining, instruction tuning operates at a much smaller computational scale; additionally, it leverages hidden representations with far lower dimensionality than gradient matrices. These advantages enable the flexible and principled use of polynomial-time algorithms to construct novel similarity-based data selection strategies.

Our contributions include:
\begin{itemize}
    \item We identify both the strengths and redundancy of representation similarity as a metric for instruction-tuning data selection.
\item We show that Rademacher projection–concatenation and classical dimensionality reduction techniques, such as whitening, are effective for semantic embedding–based instruction-tuning data selection.
\item We develop an extensible, highly parallel, and distributed data selection framework that supports end-to-end processing of millions of data, and propose an asymmetric distributed computing pattern for similarity-based problems with imbalanced candidate and test set scales.
\item On a released 16B LLM with real 2M SFT data, our experiments demonstrate the effectiveness of our algorithms: using only 3.5\% of the instruction-tuning data, our approach outperforms training on the full dataset by an average of 0.71\% across four benchmarks.

\end{itemize}

\section{Related works}
\textbf{Instruction tuning data selection.} A variety of simple, effective, and human-intuitive methods have proven highly valuable for instruction tuning data selection. Examples include perplexity \cite{DBLP:conf/iclr/YinR25, DBLP:conf/acl/AntonelloBTH20}, response length \cite{DBLP:conf/icml/ZhaoACF24}, and even random sampling \cite{DBLP:journals/corr/abs-2410-09335}. Moreover, high-quality in-context learning examples often transfer effectively to instruction tuning, owing to the substantial information gain they provide \cite{DBLP:conf/naacl/LiZLCC0W0024, DBLP:journals/corr/abs-2505-05327, DBLP:conf/acl/LiHXYYZSCLLHL24, DBLP:journals/corr/abs-2511-07074}.

Among data curation criteria, similarity has received particular attention. Gradient information is widely regarded as a powerful proxy for measuring data similarity in the context of model training, though different approaches utilize gradients in distinct ways. Some methods compute the cosine similarity between gradients induced by training and test examples to assess their alignment \cite{DBLP:conf/icml/XiaMGA024, DBLP:conf/acl/LinLX024, DBLP:conf/icml/0001L0KFL25}, while others cluster the data pool based on gradient representations \cite{DBLP:journals/corr/abs-2512-06678, DBLP:conf/acl/PanHKLLC24, DBLP:conf/naacl/ZhangQPZPZ25}. Beyond gradients, other similarity-based metrics have also demonstrated strong performance in instruction tuning data selection, including semantic representations \cite{DBLP:journals/corr/abs-2503-01807, DBLP:journals/corr/abs-2510-07118}, loss trajectories \cite{DBLP:conf/nips/0007MCM24}, influence scores \cite{DBLP:journals/corr/abs-2505-19051}, and neural tangent kernels \cite{DBLP:journals/corr/abs-2511-07380}, among others. This work falls within the line of research on semantic representation and is most closely related to RDS+ \cite{DBLP:journals/corr/abs-2503-01807}, which, to the best of our knowledge, is the only existing work dedicated to large-scale instruction tuing data selection.

Another common strategy involves employing models to rate data to reduce dataset scale. The most straightforward approach is using LLMs to directly evaluate data quality \cite{DBLP:conf/iclr/ChenLYWGYTS0HJ24}. Building on these assessments, data quality can be effectively integrated with complementary metrics—such as complexity \cite{DBLP:conf/iclr/0131Z00H24, DBLP:conf/ijcai/ZhangZLJYZLG25} and diversity \cite{DBLP:conf/emnlp/Ge0HMTZXLC0LXZ24, DBLP:conf/emnlp/BukharinLWYYLZZ24, DBLP:journals/corr/abs-2404-01067, DBLP:conf/iclr/PangW0ZW00B025}—to support more generalizable data selection. Furthermore, LLMs can be leveraged to annotate data \cite{DBLP:conf/iclr/LuY0LLTZZ24, DBLP:conf/acl/ChenLHMYC25}, and these annotations can be readily transformed into metrics to uncover relationships within the data.

\textbf{Semantic representation correction.} Some early work identified the anisotropy in representations produced by advanced encoders \cite{DBLP:conf/emnlp/Ethayarajh19}. To address this issue, a variety of methods have been proposed. Many approaches rely on contrastive learning \cite{DBLP:conf/emnlp/GaoYC21, DBLP:journals/corr/abs-2201-10005, DBLP:conf/naacl/ChuangDLZCS0YKG22}, while others focus on training normalization models \cite{DBLP:conf/emnlp/LiZHWYL20} or correcting the representations through whitening \cite{DBLP:journals/corr/abs-2103-15316}.

\section{Preliminary}
\subsection{Problem formulation}
We denote the finite vocabulary by \( \Sigma \triangleq \{w_1, w_2, \dots, w_n\} \) and define \( \Sigma^* \triangleq \bigcup_{i=0}^{\infty} \Sigma^i \), where \( \Sigma^i \) denotes the \(i\)-fold Cartesian product of \(\Sigma\) (with \(\Sigma^0 = \{\varepsilon\}\), where \(\varepsilon\) is the empty string). Without loss of generality, we consider a large language mapping \(\mathcal{\phi}(\cdot \mid \theta) : \Sigma^* \to \Sigma^*\), which is usually composed by a large language model and its post training procedure. Typically, a mature and commercially deployed language model is trained on a dataset \(X\); in this work, we refer to \(X\) as the \emph{data pool}.

To reduce the size of the data pool \(X\) while simultaneously improving the performance of the corresponding language model, we aim to select an optimal subset \(x \subseteq X\) such that the model exhibits stronger capability in solving tasks from a specified test dataset 
\begin{equation}
T \triangleq \{(t_1, a_1), (t_2, a_2), \dots, (t_m, a_m)\},
\end{equation}
where \(t_j, a_j \in \Sigma^*\) denote the \(j\)-th test input and its ground-truth answer, respectively. Formally, we seek
\begin{equation}
\argmax_{x \subseteq X} \sum_{j=1}^{|T|} \mathbb{I}\big\{ \phi_x(t_j) = a_j \big\},
\end{equation}
where \(\phi_x\) denotes the language model trained (or adapted) on the subset \(x\), and \(\mathbb{I}\{\cdot\}\) is the indicator function.

\subsection{Selecting strategy}

In this work, we select data based on representation similarity using representation engineering—specifically, by manipulating the hidden states of an LLM encoder. By convention, all representations in this paper are denoted as row vectors.
Let the encoder be a mapping
\begin{equation}
f: \Sigma^* \to \mathbb{R}^v,
\end{equation}
where \( v \) denotes the embedding dimension.
Given two embeddings \( e_1, e_2 \in \mathbb{R}^v \), the cosine similarity is defined as
\begin{equation}
\cos(e_1, e_2)
\;=\;
\frac{e_1^\top e_2}{\|e_1\|_2 \, \|e_2\|_2}
\end{equation}.
Without loss of generalization, for a test example $(t_j, a_j) \in T$, we retrieve the most similar instances from the data pool $X$ by maximizing the cosine similarity between their representations and that of the test input $t_j$. In practice, when forming a dataset at the scale $k$ for a set of test examples $\{t_1, t_2, \dots, t_m\}$, a round-robin algorithm is implemented to maintain the diversity of representations and ensure balanced coverage. 

Formally, let $S$ denote the set of selected instances, initialized as $S = \emptyset$. The selection process proceeds iteratively by cycling through the test indices $j \in \{1, \dots, m\}$. In each step, we identify the optimal candidate $x^*$ such that:
\begin{equation}
    x^* = \argmax_{x_i \in X \setminus S} \cos\big(f(t_j), f(x_i)\big)
\end{equation}
The selected instance is then appended to the set, $S \leftarrow S \cup \{x^*\}$. This cyclic procedure continues until the total number of retrieved instances reaches the target capacity, i.e., $|S| = k$. By alternating between test inputs, this strategy prevents the retrieved set from being disproportionately biased toward the nearest neighbors of a specific subset of test examples, thereby achieving a more uniform representation across the entire test distribution.

\section{Compressed representation data selection}
\subsection{A pilot study: binarized experiments}

Though semantic representations extracted by state-of-the-art encoders or LLMs exhibit strong alignment with human semantic judgments and demonstrate impressive performance in downstream applications, the underlying reasons for their effectiveness and optimization remain insufficiently understood. To this end, we design a simple binarized experiment that removes the normalization step after embedding extraction and instead retains only the sign information of each vector component.

Formally, let the representation be $e \in \mathbb{R}^v$. The corresponding binarized vector $e_b \in \mathbb{R}^v$ is defined element-wise as
\begin{equation}
    e_{b}^{(i)} \triangleq 
    \begin{cases}
        1,  & e_b^{(i)} \ge 0, \\
        -1, & e_b^{(i)} < 0 .
    \end{cases}
\end{equation}
Under this setting, we perform data selection by maximizing the vector product between binarized data-pool embeddings and original test-data embeddings (Table \ref{binarization}). The resulting selections are almost identical to those obtained using the original (non-binarized) embeddings for both the data pool and the test set.
\begin{table}[htbp]
\centering
\caption{Binarization Experiment Results. See setting in Table \ref{exp_settings}.}
\label{binarization}
\small   % 或 \footnotesize
\begin{tabular}{ccc}
\toprule
 & Original data pool & Binarized data pool \\
\midrule
Acc. on GSM & 87.64 & 87.72 \\
\bottomrule
\end{tabular}
\end{table}

\begin{figure*}[t]
  \centering
  \includegraphics[width=\textwidth]{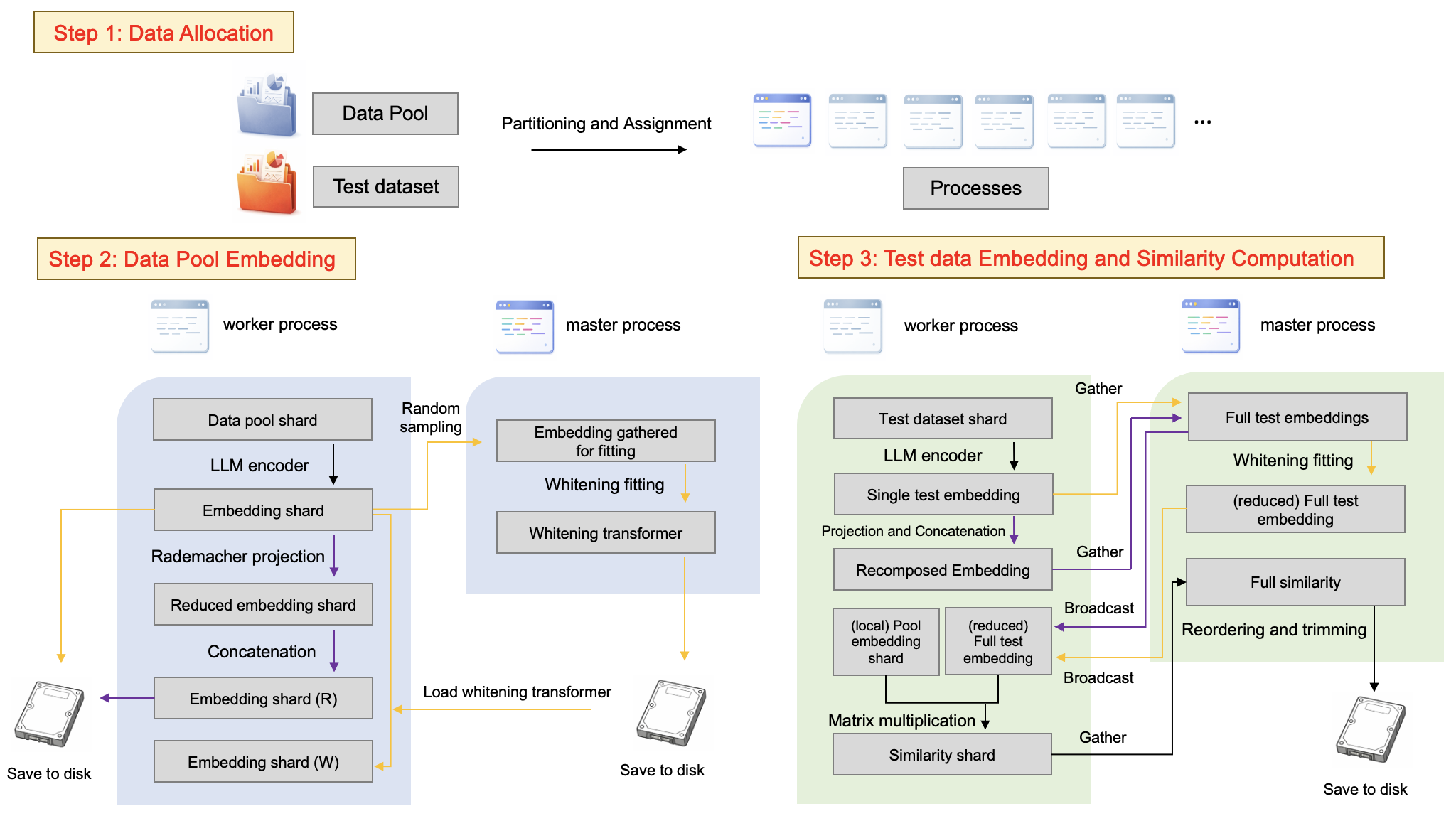}
  \caption{Overview of the proposed pipeline. Black, purple, and orange arrows denote the public flow, CRDS-R flow, and CRDS-W flow, respectively. Compared with the simpler CRDS-R, CRDS-W introduces additional gathering, fitting, and saving operations to support the whitening setting. The algorithmic details are provided in Appendix~\ref{app:alg}.}
  \label{fig:pipeline}
\end{figure*}

Moreover, we observe that when we instead use original data-pool embeddings with binarized test-data embeddings—or even binarize both embeddings—the selected datasets remain highly aligned with the GSM data type. This phenomenon strongly suggests that the current representations used for similarity computation are highly redundant, indicating substantial room for optimization. These observations motivate and inform our subsequent research.

\subsection{CRDS-R: Rademacher projection-concatenation based data selection}
Rademacher projection is a widely used technique in gradient-based data selection algorithms. This is primarily due to the enormous size of the gradient matrices involved—often matching the scale of modern large language models, which can exceed 100 billion parameters. At such scales, conventional dimensionality reduction methods like PCA become computationally infeasible. Fortunately, the Johnson–Lindenstrauss lemma guarantees that Rademacher projection preserves inner-product structures with high probability, making it especially well-suited for capturing inter-layer parameter relationships through gradient similarity—a key ingredient in effective data selection.

However, to the best of our knowledge, the impact of Rademacher projection on the preservation of \textit{semantic representations} remains unexplored. In this work, we demonstrate that Rademacher projection also performs remarkably well for semantic representations, which are typically far denser and more information-rich than sparse gradient matrices.

Let \( v \) denote the dimensionality of the hidden states produced by the encoder, and let \( w \) be the target reduced dimension. Let \( P \in \mathbb{R}^{v \times w} \) be a random projection matrix whose entries are independent and identically distributed (i.i.d.) according to the uniform distribution over \([-1, 1]\). Given an original semantic representation \( e \in \mathbb{R}^v \) extracted by the encoder, we obtain its compressed counterpart as  
\begin{equation}
e' \triangleq eP \in \mathbb{R}^w.
\end{equation}
Following the philosophy of gradient-space-based selection, the condensed representations from multiple layers are concatenated to form a new representation, which is subsequently used for cosine similarity computation. For a fair comparison, we constrain the dimensionality of the final concatenated representation to match that of the original representation.

Formally, suppose we extract \( H \) hidden-state layers from the LLM encoder. Each layer is projected to a subspace of dimension \( w\triangleq \frac{v}{H} \), such that the overall dimensionality of the concatenated representation equals the original dimensionality \( v \). To preserve layer-specific information, we employ a distinct projection matrix \( P_M \in \mathbb{R}^{v \times w} \) for each hidden-state representation \( e_M \).

The final representation used for data selection in CRDS-R is defined as
\begin{equation}
e_R
=
\begin{bmatrix}
e_1P_1 \; e_2P_2 \; \cdots \; e_HP_H
\end{bmatrix}.
\end{equation}
This construction yields a composite representation that integrates multi-layer semantic signals while maintaining computational efficiency.

% In this paper, we reveal that for some LLM encoders, their last hidden state layers contains better semantic information than the last layer, which can be easily extracted by Rademacher projection and concatenation.

\subsection{CRDS-W: Whitening-based data selection}
Whitening \cite{DBLP:journals/corr/abs-2103-15316} is a post-processing technique proposed to improve the quality of sentence embeddings by addressing the anisotropy problem commonly observed in representations from pre-trained language models. The method applies a linear whitening transformation that centers the embedding vectors (zero mean) and decorrelates their dimensions (identity covariance matrix), effectively mapping them into a standard orthogonal basis where cosine similarity becomes geometrically meaningful.

Formally, given a set of sentence embeddings $\{e_\alpha\}_{\alpha=1}^N \in \mathbb{R}^v$, the whitening algorithm first computes the mean vector 
\begin{equation}
\bar{e} = \frac{1}{N}\sum_{\alpha=1}^N e_\alpha
\end{equation}
and the covariance matrix 
\begin{equation}
C = \frac{1}{N}\sum_{i=1}^N (e_\alpha - \bar{e})^\top (e_\alpha - \bar{e}).
\end{equation}
It then performs singular value decomposition (SVD):
\begin{equation}
C = U \Lambda U^\top,
\end{equation}
where $U$ is an orthogonal matrix and $\Lambda$ is a diagonal matrix of eigenvalues. The whitening transformation matrix is constructed as $W = U \Lambda^{-1/2}$. To enable dimensionality reduction, only the first $\beta$ columns of $W$ are retained, yielding $W_\beta \in \mathbb{R}^{v \times \beta}$. Each sentence embedding is then transformed as:
\begin{equation}
    \tilde{e}_\alpha =(e_\alpha - \bar{e})W_\beta.
\end{equation}
This results in $\beta$-dimensional embeddings that are both isotropic and compact. The reduced dimensionality not only enhances semantic retrieval performance in many cases but also significantly lowers storage requirements and accelerates similarity search. 

% Preamble (usually already included)

% In the paper body (place near where you first mention it)

\begin{table*}[t]
\centering
\caption{Experimental results, selecting 70k data from Ling 2.0 (2M data).
The base model is Ling-mini-2.0, using a truncation length of 2048.
Each reported value represents the average of five parallel runs. }
\label{main70k}
\setlength{\tabcolsep}{12pt}
\begin{tabular}{ccccccc}
\toprule
Model & Method & GSM & MMLU & MBPP & BBH & Average \\
\midrule
\multirow{7}{*}{\parbox[c]{2cm}{\centering Ling-mini-2.0\\(16B)}}
& Ling-mini-2.0 (100\%) & 90.75 & \textbf{78.50} & \underline{84.07} & 79.68 &  83.25      \\
& Random (3.5\%)         & 91.28 & 77.28 & 83.37 & 78.39 &  82.58      \\
& Mid PPL (3.5\%)        & 91.22 & 76.21 & 83.28 & 76.76 &   81.87     \\
& Length (3.5\%)         & 89.02 & 75.35 & 74.85 & 57.74 &  74.24      \\
& RDS+ (3.5\%)           & 91.10 & 77.84 & 82.58 & 78.87 &   82.60     \\
& CRDS-R (3.5\%)    & \underline{91.40} & 77.36 & 83.75 & \underline{80.54} & \underline{83.26}       \\
& CRDS-W (3.5\%)    & \textbf{92.55} & \underline{77.99} & \textbf{84.26} & \textbf{81.04} &  \textbf{83.96}      \\
\bottomrule
\end{tabular}
\end{table*}

\begin{table*}[t]
\centering
\caption{Experimental results, selecting 100k data from Ling 2.0 (2M data). The base model is Ling-mini-2.0, using a truncation length of 2048.
Each reported value represents the average of five parallel runs.}
\label{main100k}
\setlength{\tabcolsep}{12pt}
\begin{tabular}{ccccccc}
\toprule
Model & Method & GSM & MMLU & MBPP & BBH & Average \\
\midrule
\multirow{7}{*}{\parbox[c]{2cm}{\centering Ling-mini-2.0\\(16B)}}
& Ling-mini-2.0 (100\%) & 90.75 & \textbf{78.50} & 84.07 & 79.68 & 83.25       \\
& Random (5\%)         & \underline{91.48} & 76.44 & 83.28 & 79.01 &  82.55      \\
& Mid PPL (5\%)         & 91.01 & 76.50 & 83.98 & 78.48 &   82.49     \\
& Length (5\%)          & 85.05 & 74.77 & 78.08 & 60.22 &   74.53     \\
& RDS+ (5\%)            & 90.55 & 76.44 & 83.42 & 78.72 &   82.28     \\
& CRDS-R (5\%)     & 91.02 & 77.54 & \underline{84.26} & \textbf{80.44} &    \underline{83.31}    \\
& CRDS-W (5\%)     & \textbf{92.16} & \underline{77.99} & \textbf{85.20} & \underline{80.10} &    \textbf{83.86}    \\
\bottomrule
\end{tabular}
\vspace{-0.2cm}
\end{table*}

\subsection{Distributed data selection framework}
To develop an effective and practical data selection framework for instruction tuning, we build our system on top of RDS+ with several key modifications. A major limitation of the original framework is its monolithic processing of the entire dataset, which requires both the data pool and the test set to be fully loaded into memory. More critically, peak memory usage occurs when embeddings for the data pool and the test set are stored simultaneously. This often exceeds the memory capacity of standard cloud GPU nodes, rendering the framework impractical for real-world deployment.

To address these limitations, we propose a highly parallelized, end-to-end, and memory-efficient framework for data selection. For the data pool, we adopt a partition–store–reconstruct strategy. The data are interleaved and partitioned across GPUs, where embeddings are computed in a distributed manner. Each GPU then writes its embedding shard to disk for reuse without recomputation. For the typically smaller test dataset, embeddings are computed in parallel and gathered by the main process to form the complete test embedding matrix, which is broadcast to all workers. Each GPU computes a local similarity matrix between its data-pool shard and the full test embeddings. Finally, the main process aggregates all partial similarity matrices into the complete result and saves it to disk.

Overall, our framework fully exploits multi-GPU computing resources while remaining memory efficient. By explicitly leveraging the scale asymmetry between the data pool and the test dataset, we deliver a high-quality, end-to-end, and deployable similarity computation framework that is both effective and practical for real-world instruction tuning pipelines. Figure \ref{fig:pipeline} and Algorithm \ref{alg:dds} illustrate the overall idea of our framework.

\section{Experiments}
\subsection{Data pools}
\textbf{Ling 2.0 \cite{DBLP:journals/corr/abs-2510-22115}.} We use the official release version of Ling 2.0's fine-tuning data, which includes 2.0M items. This is a balanced SFT dataset that integrates reasoning, general, and industrial tasks within a dual-mode prompt framework. It spans math, logic, coding, creativity, dialogue, and domain-specific workflows across finance, healthcare, and operations. 

\textbf{Ling 2.0 Dev.} This is a development version of the Ling 2.0 SFT dataset, containing 1.47M high-quality SFT samples with a distribution highly similar to that of Ling 2.0. This dataset is used for most hyperparameter tuning and exploratory experiments.

\subsection{Baselines} We consider the following representative baselines as our comparison groups:

\textbf{Random Selection.} Random selection serves as a strong and essential baseline for instruction tuning data selection. Under this approach, all samples in the dataset are treated equally, and a subset of a specified size is selected uniformly at random to form the experimental group.
\begin{figure*}[t]
  \centering
  \begin{subfigure}[b]{0.32\textwidth}
    \centering
    \includegraphics[width=\textwidth]{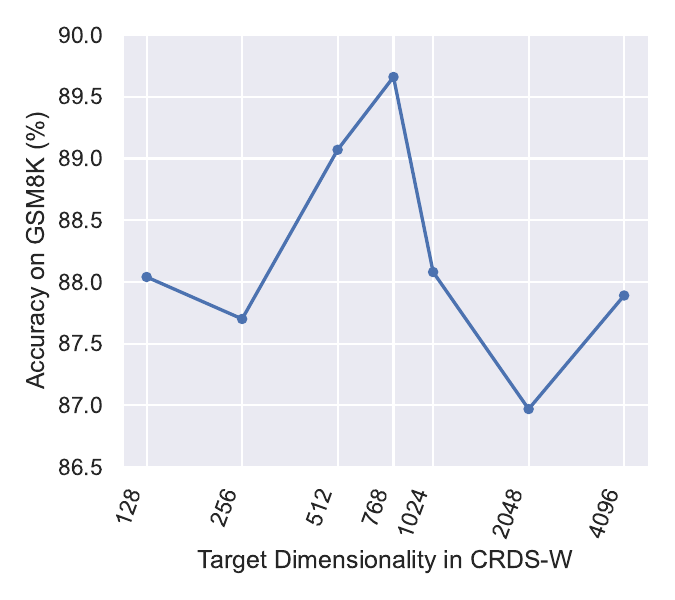}
    \caption{The effect of whitening target dimensionality.}
    \label{whiten_ab}
  \end{subfigure}
  \hfill
  \begin{subfigure}[b]{0.32\textwidth}
    \centering
    \includegraphics[width=\textwidth]{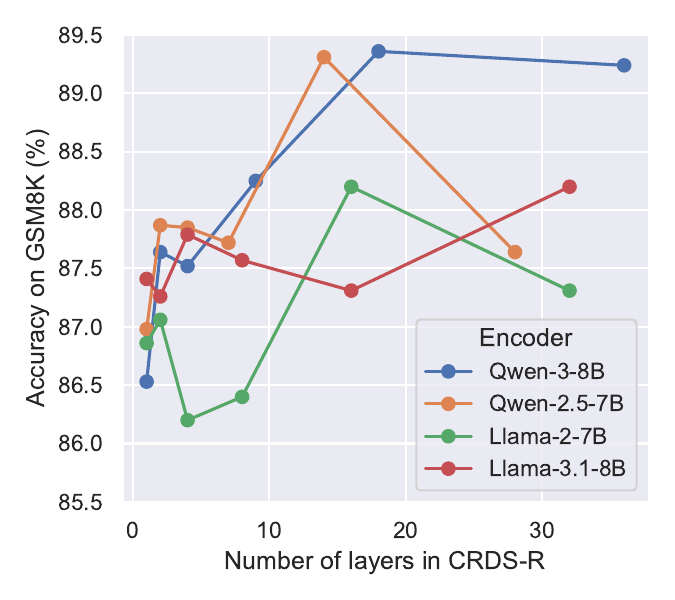}
    \caption{The effect of number of hidden states $H$ used in CRDS-R.}
    \label{layer_ab}
  \end{subfigure}
  \hfill
  \begin{subfigure}[b]{0.32\textwidth}
    \centering
    \includegraphics[width=\textwidth]{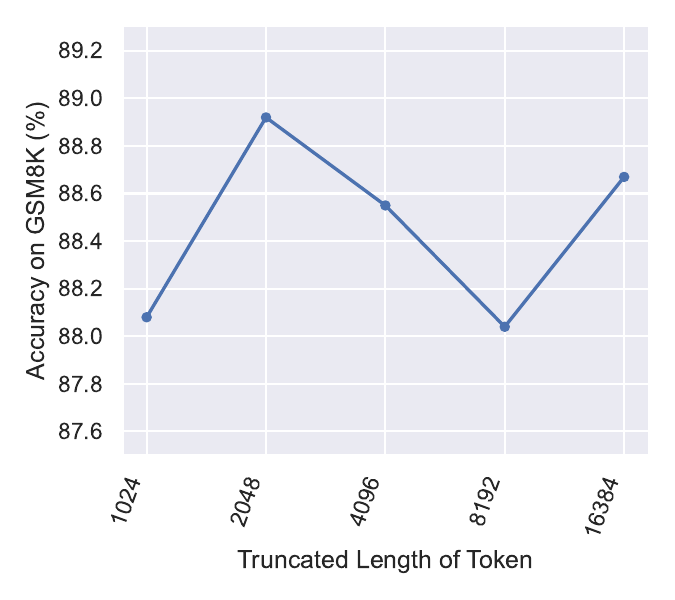}
    \caption{The effect of truncation length of LLM encoders.}
    \label{trunc_ab}
  \end{subfigure}
  \caption{Results of the ablation studies are presented in Section \ref{ablation}. See settings in Table \ref{exp_settings}.}
  \label{fig:three_side_by_side}
  \vspace{-0.2cm}
\end{figure*}

\textbf {Data Length.} Response length provides an intuitive proxy for data quality. Following \citet{DBLP:conf/icml/ZhaoACF24}, we use the total response length as a baseline metric. For multi-turn dialogues, all responses are concatenated, and their combined length is used as the scoring criterion.

\textbf{Perplexity.} Perplexity is a widely used metric for evaluating language models. It quantifies the model’s uncertainty in predicting the next token in a sequence: lower perplexity indicates stronger predictive performance, reflecting a better grasp of the underlying language distribution and an ability to generate more fluent and natural text. In this work, we adopt Mid-PPL \cite{DBLP:conf/iclr/YinR25} as our baseline, selecting data corresponding to the 30th to 60th percentiles of the perplexity distribution.

\textbf{Semantic Representation Similarity.} Our work aligns with the line of research based on representation similarity. Accordingly, we adopt the state-of-the-art method RDS+ \cite{DBLP:journals/corr/abs-2503-01807} as the baseline of representation-based data selection. RDS+ uses the Llama-2-7B model as an encoder and extracts the hidden states from the last transformer layer to serve as the semantic representation of each data instance.

\subsection{CRDS algorithms}
\textbf{CRDS-R} integrates seamlessly into the forward computation process. Prior to the forward pass, we initialize a set of $H$ Rademacher projection matrices, each corresponding to one of the $H$ hidden layers to be utilized. After feeding the input through the encoder and completing the forward pass, the hidden-state representations from the selected layers are extracted simultaneously. Each representation is then projected using its associated matrix to yield a $v/H$-dimensional vector. These projected vectors are concatenated to form a compact representation whose dimensionality matches that of the original representation space. In practice, we find that reducing and concatenating the hidden states from the final layers works effectively for data selection. For instance, in our main experiments with the Qwen-3-8B encoder, we use the last 18 layers. Moreover, systematic sampling of hidden states at regular intervals also proves effective in certain scenarios, demonstrating the flexibility of the representation design (Appendix \ref{syssampling}). Algorithm~\ref{alg:crdsr} presents the core procedure of CRDS-R.

\textbf{CRDS-W} adapts the standard whitening algorithm to our distributed computing framework. Initially, embeddings are computed in parallel and partitioned into shards across $D$ worker processes. Let $F$ denote the number of samples used to estimate the whitening transformation. Each process contributes $F/D$ embeddings to the main process, which aggregates them to compute the whitening matrix. This matrix is then saved to disk and broadcast to all processes to transform the remaining pool and test embeddings. A notable advantage of this approach is that the whitening matrix, once computed, can be reused across different test datasets, enhancing computational efficiency in multi-dataset evaluation scenarios. In our main experiments, we set $F = 500{,}000$ and $D = 8$, and select the best-performing configuration between output dimensionalities $\beta = 512$ and $\beta = 1024$. All whitening experiments use Llama-2-7B as the encoder. Algorithm~\ref{alg:crdsw} outlines the complete procedure of CRDS-W.

\subsection{Experimental results}
Tables \ref{main70k} and \ref{main100k} report our main experimental results.

\textbf{CRDS achieves the best average performance and outperforms the full dataset.}  
For the Ling-mini-2.0 base model, we sample 70K and 100K examples from the original 2M data pool. Under both selection budgets, CRDS-W attains the highest average accuracy across four benchmarks, surpassing the full-data baseline by 0.71\% and 0.61\% at 70K and 100K, respectively. CRDS-R also exceeds the full baseline, yielding gains of 0.01\% at 70K and 0.06\% at 100K. These results demonstrate that our dimensionality-reduction-based data selection methods can match or even outperform the full-dataset baseline while using only 3.5\%--5\% of the training data.

\textbf{CRDS decisively outperforms the SOTA representation-based method RDS+.}  
Compared with RDS+, which relies on the encoder's last hidden layer as the semantic representation, both CRDS variants exhibit clear advantages. Specifically, CRDS-W outperforms RDS+ by 1.36\% and 1.54\% at the 70K and 100K scales, respectively, while CRDS-R surpasses RDS+ by 0.66\% and 1.03\%. These improvements indicate that our proposed representations are better suited for instruction-tuning data selection. At the dataset level, CRDS-W exceeds the full-data baseline on GSM, MBPP, and BBH, highlighting its potential for general-domain tasks. CRDS-R shows a similar trend, albeit with slightly weaker overall performance than CRDS-W. Nonetheless, both CRDS methods consistently outperform RDS+: CRDS-R surpasses RDS+ in 7 out of 8 settings, while CRDS-W achieves superior performance in all 8 cases.

\section{Ablation Study} \label{ablation}

\subsection{Raw performance of encoders} Following the experimental setup of \citet{DBLP:journals/corr/abs-2503-01807}, we adopt Llama-2-7B as the starting-point encoder. Our results in Table \ref{raw_encoder} corroborate the finding reported by RDS+ \cite{DBLP:journals/corr/abs-2503-01807}: newer encoder models do not necessarily yield superior performance in data selection tasks. 
Notably, four different encoders perform very similarly on the data selection task, underscoring that model scale and recency alone are insufficient indicators of encoder effectiveness for this task.

\begin{table}[ht]
\centering
\caption{Raw encoder performance. See settings in Table \ref{exp_settings}.}
\label{raw_encoder}
\vskip 0.15in
\resizebox{\columnwidth}{!}{%
\begin{tabular}{ccccc}
\toprule
Model & Qwen-2.5-7B & Qwen-3-8B & Llama-2-7B & Llama-3.1-8B \\
\midrule
Accuracy on GSM8K (\%) & 87.59 & 87.69 & 87.64 & 87.44 \\
\bottomrule
\end{tabular}
}
\vskip -0.1in
\end{table}

\subsection{Selection of truncation length} Truncated sequence length is a critical factor that directly affects how much informative content is retained during data selection. Intuitively, a longer context window should encode richer and more informative representations; however, the results in Table \ref{trunc_ab} show that this intuition does not hold for data selection tasks. For Qwen-2.5-7B, when applying the CRDS-R algorithm with the last 14 layers selected, a context window of 2048 tokens emerges as the optimal choice. This may be because, in most cases, 2048 tokens are sufficient to capture the core ideas of a paragraph, or because truncation during SFT fundamentally influences the effectiveness of data selection. Accordingly, we adopt a truncation length of 2048 tokens for the experiments reported in Tables \ref{main70k} and \ref{main100k}.

\subsection{Selection of whitening dimensionality}

The whitening procedure is sensitive to the choice of final dimensionality. As noted by \citet{DBLP:journals/corr/abs-2103-15316}, the effective dimensionality of a representation can vary substantially across tasks. Motivated by this observation, we conduct a systematic study of how the output dimensionality affects whitening performance. In all experiments, we use Llama-2-7B as the encoder, which has a hidden state size of 4096, and sweep the target dimensionality $\beta$ over \{128, 256, 512, 768, 1024, 2048, 4096\}. In Table \ref{whiten_ab}, we see performance peaks at 768 dimensions, suggesting that retaining approximately 20\% of the original dimensionality is an effective choice for whitening. In our main experiments, we evaluated target dimensionalities of 512 and 1024 and report the best-performing results (Appendix \ref{CRDSW-best}).

\subsection{Effect of hidden layers in CRDS-R} 
We conduct controlled experiments using four language models as encoders:
Qwen-2.5-7B, Qwen-3-8B, Llama-2-7B, and Llama-3.1-8B.
In this section, we investigate the effect of varying the number of hidden states of LLM transformer. Specifically, we set the number of hidden states to
$H \in \{1, 2, 4, 7, 14, 28\}$ for the Qwen-2.5-7B encoder,
$H \in \{1, 2, 4, 9, 18, 36\}$ for the Qwen-3-8B encoder, and
$H \in \{1, 2, 4, 8, 16, 32\}$ for the Llama-based encoders. The results in Table~\ref{layer_ab} show that CRDS-R is sensitive to the choice of extracted layers. In particular, the Qwen encoders benefit more from concatenating multiple layers: Qwen-2.5-7B and Qwen-3-8B achieve their best performance when $H=14$ and $H=18$, respectively. In contrast, the performance gains from CRDS-R are less pronounced for the Llama models. Additionally, our preliminary experiments show that Qwen-2.5-7B performs worse than Qwen-3-8B (Appendix \ref{CRDSR-encoder}). Based on these observations, we adopt Qwen-3-8B with $H=18$ as the default setting in Table~\ref{main70k} and Table~\ref{main100k}.

\section{Conclusion}
In this work, we systematically investigate the redundancy of semantic representations used in instruction-tuning data selection. We propose a distributed computing strategy that accommodates asymmetric scales between candidate and test data, and develop an effective framework for instruction-tuning data selection. Upon identifying redundancy in existing representations, we introduce two methods—CRDS-R and CRDS-W—based on Rademacher projection–concatenation and whitening, respectively, to optimize semantic representations for similarity computation. Our approach consistently outperforms both full-data baselines and state-of-the-art methods on industrial-scale LLMs and corresponding large-scale SFT datasets.

Despite strong performance under the current setting, the generalization of our methods to smaller base models remains an open question, as such models behave differently from large LLMs. Moreover, the mechanism underlying representation-similarity–based data selection is not yet well understood. It is also unclear whether gradient-based or representation-based approaches are more effective for data selection. We hope this work will inspire further investigation of data selection methods within the community.


% Authors are kindly asked to make their submissions as accessible as possible
% for everyone including people with disabilities and sensory or neurological
% differences. Tips of how to achieve this and what to pay attention to will be
% provided on the conference website \url{http://icml.cc/}.

% \section*{Software and Data}

% If a paper is accepted, we strongly encourage the publication of software and
% data with the camera-ready version of the paper whenever appropriate. This can
% be done by including a URL in the camera-ready copy. However, \textbf{do not}
% include URLs that reveal your institution or identity in your submission for
% review. Instead, provide an anonymous URL or upload the material as
% ``Supplementary Material'' into the OpenReview reviewing system. Note that
% reviewers are not required to look at this material when writing their review.

% % Acknowledgements should only appear in the accepted version.
\section*{Acknowledgements}
We sincerely thank Professor Bo Li for her valuable support and assistance in this project.

% \textbf{Do not} include acknowledgements in the initial version of the paper

% If a paper is accepted, the final camera-ready version can (and usually should)
% include acknowledgements.  Such acknowledgements should be placed at the end of
% the section, in an unnumbered section that does not count towards the paper
% page limit. Typically, this will include thanks to reviewers who gave useful
% comments, to colleagues who contributed to the ideas, and to funding agencies
% and corporate sponsors that provided financial support.

\section*{Impact Statement}

This work does not involve any activities that pose potential harm to individuals, communities, or society. It does not include direct or indirect interventions involving human participants, animals, or the environment, nor does it involve the collection or processing of sensitive, private, or personally identifiable information. The work is conducted in accordance with applicable laws, regulations, and ethical guidelines, and its outcomes are intended solely for legitimate academic or practical purposes. Therefore, no foreseeable risks or negative impacts are expected to arise from this work.


\bibliography{example_paper}
\bibliographystyle{icml2026}

%%%%%%%%%%%%%%%%%%%%%%%%%%%%%%%%%%%%%%%%%%%%%%%%%%%%%%%%%%%%%%%%%%%%%%%%%%%%%%%
%%%%%%%%%%%%%%%%%%%%%%%%%%%%%%%%%%%%%%%%%%%%%%%%%%%%%%%%%%%%%%%%%%%%%%%%%%%%%%%
% APPENDIX
%%%%%%%%%%%%%%%%%%%%%%%%%%%%%%%%%%%%%%%%%%%%%%%%%%%%%%%%%%%%%%%%%%%%%%%%%%%%%%%
%%%%%%%%%%%%%%%%%%%%%%%%%%%%%%%%%%%%%%%%%%%%%%%%%%%%%%%%%%%%%%%%%%%%%%%%%%%%%%%
\newpage
\appendix
\onecolumn
\section{Algorithms} \label{app:alg}
We present the principal algorithms used and/or proposed in this paper. 

\textbf{Algorithm \ref{alg:dds}} describes the parallelized representation-based selection framework. It presents the core procedures for representation extraction and similarity computation (Figure \ref{fig:pipeline}), without incorporating CRDS-R or CRDS-W. After initialization, the data pool and test set are distributed to all worker processes. The data pool is first embedded and stored, followed by embedding the test data. Because the test set is much smaller, we adopt an asymmetric strategy: test embeddings are gathered and multiplied with sharded data pool embeddings. Finally, we gather and store the similarity shards, which are orders of magnitude smaller than the original representations. 

\textbf{Algorithm \ref{alg:crdsr}} illustrates the core idea of the CRDS-R method, which extracts multiple hidden states from the transformer in a single forward pass, followed by Rademacher projection and concatenation of the extracted representations. 

\textbf{Algorithm \ref{alg:crdsw}} emphasizes its differences from Algorithm \ref{alg:dds}, introducing additional fitting, gathering, saving, and loading procedures. In Algorithm \ref{alg:crdsw}, the grey-colored lines indicate the steps that are identical to those in Algorithm \ref{alg:dds}.
\vspace{-0.2cm}
\begin{figure*}[htbp]
\centering
\begin{minipage}[t]{0.48\textwidth}
    \begin{algorithm}[H]
        \caption{Distributed Data Similarity Computation}
        \label{alg:dds}
        \begin{algorithmic}[1]
            \REQUIRE Data pool $X$, test dataset $T$, encoder $f$, number of GPUs $n$
            \ENSURE Full similarity matrix $S$

            \STATE \textbf{Initialize processes and partition data}
            \STATE $i \leftarrow \text{rank}()$ \COMMENT{$i \in \{0,\dots,n-1\}$}
            \STATE $(X_0,\dots,X_{n-1}) \leftarrow \text{InterleavedSplit}(X,n)$
            \STATE $(T_0,\dots,T_{n-1}) \leftarrow \text{InterleavedSplit}(T,n)$ 
            \STATE $X_i, T_i$ are local to process $i$

            \STATE
            \STATE \textbf{Encode data pool (local)}
            \STATE $e_{Xi} \leftarrow \emptyset$
            \FOR{$x \in X_i$}
              \STATE $e_{Xi} \leftarrow e_{Xi} \cup \{f(x)\}$
            \ENDFOR
            \STATE Save $e_{Xi}$ to disk 

            \STATE
            \STATE \textbf{Encode test data (local)}
            \STATE $e_{Ti} \leftarrow \emptyset$
            \FOR{$t \in T_i$}
              \STATE $e_{Ti} \leftarrow e_{Ti} \cup \{f(t)\}$
            \ENDFOR

            \STATE
            \STATE \textbf{Collect and broadcast test embeddings}
            \IF{$i = 0$}
              \STATE $e_T \leftarrow \text{Gather}(e_{T0},\dots,e_{T(n-1)})$ 
            \ENDIF
            \STATE $e_T \leftarrow \text{Broadcast}(e_T)$

            \STATE
            \STATE \textbf{Compute similarity (local)}
            \STATE $S_i \leftarrow e_{Xi} \cdot e_T^\top$

            \STATE
            \STATE \textbf{Assemble results (main)}
            \IF{$i = 0$}
              \STATE $S \leftarrow \text{Gather}(S_0,\dots,S_{n-1})$
              \STATE $S \leftarrow \text{RearrangeAndTrimPadding}(S)$
              \STATE Save $S$ to disk
            \ENDIF
        \end{algorithmic}
    \end{algorithm}
\end{minipage}
\hfill
\begin{minipage}[t]{0.48\textwidth}
    \begin{algorithm}[H]
        \caption{CRDS-R}
        \label{alg:crdsr}
        \begin{algorithmic}[1]
            \REQUIRE Projection matrices $\{P_h\}_{h=1}^H$, encoder $f$, data $x$
            \ENSURE CRDS-R representation $e_R$
            \STATE $e_1, e_2, \cdots, e_H \leftarrow f(x)$
            \STATE $e'_{1}, e'_{2}, \cdots, e'_H \leftarrow e_1P_1, e_2P_2, \cdots, e_HP_H $ 
            \STATE $e_R \leftarrow \textrm{Concat}(e'_1, e'_2, \cdots, e'_H)$ 
        \end{algorithmic}
    \end{algorithm}

    \vspace{1.2em} 

    \begin{algorithm}[H]
    \caption{CRDS-W}
    \label{alg:crdsw}
    \begin{algorithmic}[1]
        \REQUIRE Data pool $X$, test dataset $T$, target dimensionality $\beta$, encoder $f$, number of fitting examples $F$, number of GPUs $n$
        \ENSURE Full similarity matrix $S$

        \STATE {\color{gray}\textbf{Initialize processes and partition data}} 
        \STATE {\color{gray}\textbf{Encode data pool (local)}}
        \STATE
        \STATE \textbf{Random sampling (local)}
        \STATE $\hat{e}_{i}\leftarrow$ RandomSample($e_{Xi}$, $F/n$) 
        \STATE
        \STATE \textbf{Gather and fit (main)}
        \STATE $\hat{e} \leftarrow \textrm{Gather}(\hat{e}_{0}, \dots, \hat{e}_{n-1})$
        \STATE Whitening transformer $W \leftarrow \textrm{WhiteningFit}(\hat{e}, \beta)$
        \STATE Save $W$ to disk 
        \STATE
        \STATE \textbf{Whiten data pool embeddings (local)}
        \STATE Load $W$ from disk
        \STATE $\tilde{e}_{Xi} \leftarrow W(e_{Xi})$
        \STATE 
        \STATE {\color{gray}\textbf{Encode test data (local)}}
        \STATE {\color{gray}\textbf{Collect and broadcast test embeddings}}
        \STATE
        \STATE \textbf{Whiten test data and compute (local)}
        \STATE $\tilde{e}_{T} \leftarrow W(e_{T})$
        \STATE $S_i \leftarrow \tilde{e}_{Xi} \cdot \tilde{e}_T^\top$
        \STATE
        \STATE {\color{gray}\textbf{Assemble results (main)}}
    \end{algorithmic}
\end{algorithm}
\end{minipage}

\end{figure*}

\section{Supplementary Experimental Settings}

\subsection{Benchmarks and Test Datasets.} 
Similar to the settings in RDS+ \cite{DBLP:journals/corr/abs-2503-01807}, we use a small split of the test dataset within the computing framework. For ease of evaluation and to avoid potential data leakage, we only use datasets that naturally provide a development or validation split, or official few-shot examples. Below, we list all the datasets used in our experiments, and Table~\ref{test_ds_detail} summarizes the details of the test examples employed in the computing framework. When evaluating the model performance, we use the test split of the dataset within the OpenCompass framework.

\textbf{GSM8K} \cite{DBLP:journals/corr/abs-2110-14168} is a dataset of 8,500 high-quality, multi-step grade-school math word problems created by human experts. It evaluates a model’s ability to perform complex reasoning through step-by-step solutions. Designed for chain-of-thought prompting, it serves as a benchmark for arithmetic and logical reasoning in language models. 

\textbf{MMLU} \cite{DBLP:conf/iclr/HendrycksBBZMSS21} covers 57 subjects across STEM, humanities, social sciences, and more. It tests knowledge and reasoning at high school and professional levels. With 15,908 questions, it evaluates broad factual understanding and zero-shot generalization, making it a comprehensive benchmark for large language models’ academic proficiency.

\textbf{MBPP} \cite{DBLP:journals/corr/abs-2108-07732} contains 974 short Python programming tasks written by novice programmers. Each problem includes a natural language description and test cases. It assesses code generation ability in real-world beginner scenarios, focusing on correctness, simplicity, and functional implementation without requiring advanced algorithms or external libraries. To obtain more informative and reliable results, we used the sanitized MBPP to evaluate the model performance.

\textbf{BBH} \cite{DBLP:conf/acl/SuzgunSSGTCCLCZ23} is a curated subset of 23 challenging tasks from the BIG-Bench benchmark where prior models scored below random chance. It targets complex reasoning—such as logical deduction, mathematics, and linguistics—and serves as a high-difficulty benchmark to evaluate advanced cognitive capabilities in state-of-the-art language models.

\begin{table}[htbp]
\centering
\caption{Test datasets used for computing embeddings}
\label{test_ds_detail}
\begin{tabular}{ccc}
\toprule
Test dataset & Computing subset           & Data scale \\
\midrule
GSM8K        & Official few-shot examples & 8          \\
MMLU         & Official dev set           & 285        \\
MBPP         & Official prompt set        & 7          \\
BBH          & Official few-shot examples & 81         \\
\bottomrule
\end{tabular}
\end{table}

\subsection{Encoders and Base Models.}
\textbf{Encoder.} In our work, Qwen-3-8B \cite{DBLP:journals/corr/abs-2505-09388}, Qwen-2.5-7B \cite{DBLP:journals/corr/abs-2412-15115}, Llama-2-7B \cite{DBLP:journals/corr/abs-2307-09288}, and Llama-3.1-8B \cite{DBLP:journals/corr/abs-2407-21783} are evaluated in preliminary experiments to identify the most suitable model for the majority of our experiments. All of these models are LLM encoders with slightly different architectures. The main parameters of each model are summarized in Table \ref{tab:llm-arch}.

\begin{table}[htbp]
\centering
\caption{Architecture of the evaluated LLM encoders.}
\label{tab:llm-arch}
\begin{tabular}{ccc}
\toprule
LLM encoder & Hidden layers & Hidden dim. \\
\midrule
Qwen-2.5-7B  & 28 & 3584 \\
Qwen-3-8B    & 36 & 4096 \\
Llama-2-7B   & 32 & 4096 \\
Llama-3.1-8B & 32 & 4096 \\
\bottomrule
\end{tabular}
\end{table}

\textbf{Base models.} We adopt Ling-2.0-mini as the base model for our main experiments and associated ablations, while Qwen-2.5-7B serves as the base model for experiments involving smaller-scale architectures.

\subsection{Training and Evaluation Settings.} 

% We employ two distinct training pipelines for Ling-Mini model and Qwen-2.5-7B model. 

The Ling-mini-2.0 base model is trained on one of our corporation’s internal training platforms. Upon completion, the models are transferred to an internal evaluation platform largely built upon OpenCompass. For the Ling 2.0 base model, they are trained on 8-32 H200 or H800 GPUs for 3 epochs, following the routine parameter setting of Ling base models. Due to the instability of the training process, the results exhibit noticeable fluctuations, making them somewhat difficult to interpret. To improve the reliability and validity of our work, most experiments reported in this paper are repeated three or five times, and the results are averaged to form a single data point. 

\section{Supplementary Experimental Results}
\subsection{Experimental Results of CRDS-R for Different Encoders} \label{CRDSR-encoder}
As shown in Table~\ref{layer_ab}, both Qwen-2.5-7B and Qwen-3-8B achieve better performance than the setting without CRDS-R, and both outperform the Llama-2-7B baseline.
To determine which encoder performs better under our primary experimental settings, we conduct additional experiments, setting $H=14$ for Qwen-2.5-7B and $H=18$ for Qwen-3-8B. The results are reported in Table~\ref{CRDSR_sup}.
Based on these results, we select Qwen-3-8B with $H=18$ as the default configuration in Table~\ref{main70k} and Table~\ref{main100k}.
 
\begin{table}[htbp]
\centering
\caption{Performance under different encoders for CRDS-R.}
\label{CRDSR_sup}
\begin{tabular}{ccccc}
\toprule
\multirow{2}{*}{Encoder} & \multicolumn{4}{c}{Dataset} \\
\cmidrule(lr){2-5}
                         & GSM8K (70k) & MBPP (70k) & GSM8K (100k) & MBPP (100k) \\
\midrule
Qwen-2.5-7B, $H=14$             & 91.31       & 83.84      & 90.11        & 83.04       \\
Qwen-3-8B, $H=18$                & 91.40       & 83.75      & 91.02        & 84.26       \\
\bottomrule
\end{tabular}
\end{table}

\subsection{Experimental Results of CRDS-R for Systematic Sampling of Hidden States} \label{syssampling}
In addition to the experimental results showing that extracting the continuous last hidden states of LLM encoders yields strong performance on CRDS-R, we also find that systematically sampling hidden states across layers performs well, even when the encoder is warmed up. However, there are limitations in layer selection. For example, the final layer should be retained: when we experimented with using only the middle layers of the encoder, the model could not be trained successfully.
\begin{figure}[htbp]
    \centering
    \begin{subfigure}{0.48\linewidth}
        \centering
        \includegraphics[width=\linewidth]{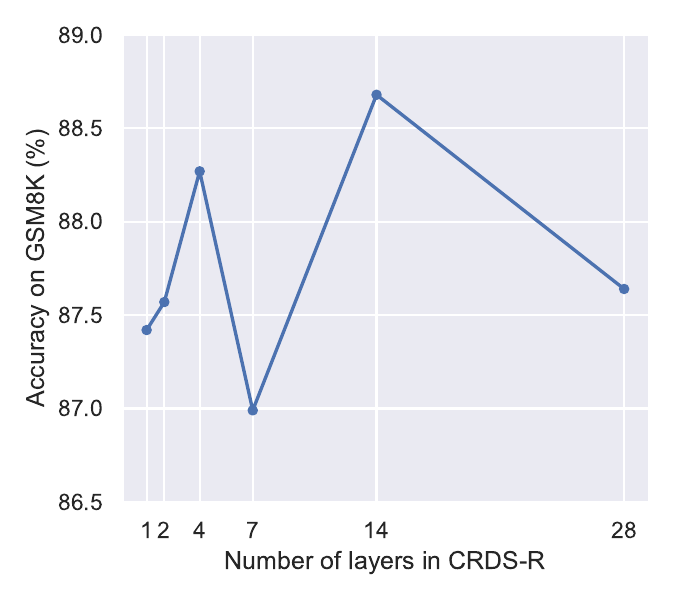}
        \caption{Qwen-2.5-7B encoder.}
        \label{fig:multilayer1}
    \end{subfigure}
    \hfill
    \begin{subfigure}{0.48\linewidth}
        \centering
        \includegraphics[width=\linewidth]{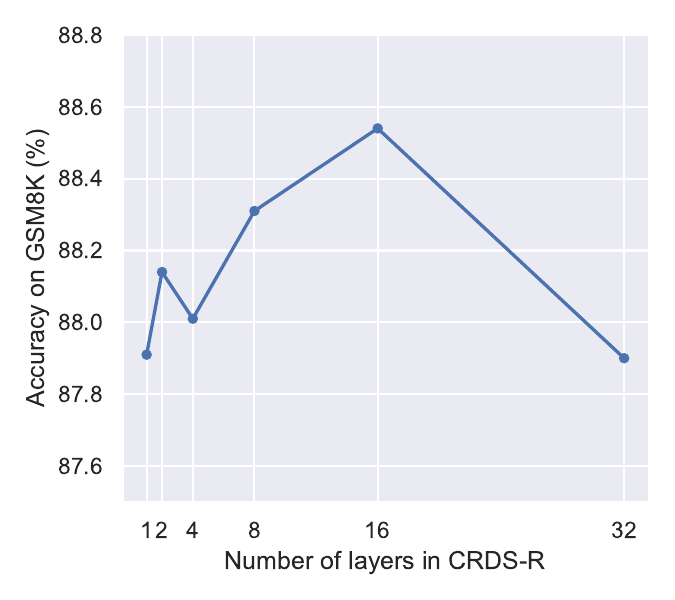}
        \caption{Llama-2-7B encoder warmed up on 100K samples from the Ling-2.0-dev dataset.}
        \label{fig:multilayer2}
    \end{subfigure}
    \caption{Effect of $H$ on systematic extraction in CRDS-R. See settings in Table \ref{exp_settings}.}
    \label{fig:CRDSR-multilayer}
\end{figure}

\subsection{Experimental Results of CRDS-W for Differenct $\beta$} \label{CRDSW-best}
For the CRDS-W experiments reported in Tables \ref{main70k} and \ref{main100k}, we evaluated two target whitening dimensionalities, $\beta = 512$ and $\beta = 1024$. The corresponding results are presented in Tables \ref{CRDSW70k} and \ref{CRDSW100k}. For each setting, we report the best performance obtained across the two dimensions.

\begin{table}[htbp]
\centering
\caption{Performance under Different Target Dimensionalities, 70k in 2M.}
\label{CRDSW70k}
\begin{tabular}{ccccc}
\toprule
\multirow{2}{*}{$\beta$} & \multicolumn{4}{c}{Datasets} \\
\cmidrule(lr){2-5}
 & GSM & MMLU & MBPP & BBH \\
\midrule
512  & 92.55 & 77.99 &84.26  &78.28  \\
1024 & 91.81 & 77.33 &81.69  &81.04  \\
\bottomrule
\end{tabular}
\end{table}

\begin{table}[htbp]
\centering
\caption{Performance under Different Target Dimensionalities, 100k in 2M.}
\label{CRDSW100k}
\begin{tabular}{ccccc}
\toprule
\multirow{2}{*}{$\beta$} & \multicolumn{4}{c}{Datasets} \\
\cmidrule(lr){2-5}
 & GSM & MMLU & MBPP & BBH \\
\midrule
512  & 92.16 & 77.99 &85.20  &72.63  \\
1024 & 91.98 &77.75  &82.81  &80.10  \\
\bottomrule
\end{tabular}
\end{table}

\section{Computational Overhead}
Both CRDS-R and CRDS-W introduce only modest additional computational overhead to the system. For CRDS-R, when all 32 layers of Llama-2-7B are selected, and the data pool consists of 1.47M samples from the Ling 2.0 development set, the total computation time increases by only 4\% compared to the original RDS+, rising from 9,237 seconds to 9,605 seconds.

In contrast, CRDS-W relies on a more complex whitening procedure, whose computational cost depends on both the fitting scale and the target dimensionality. Under the same experimental setting as CRDS-R, with an additional fitting set of 500,000 samples, we report the fitting time for different target dimensionalities in Table~\ref{tab:fitting_time}. Based on these results, and considering the target dimensionality used for Ling~2.0, the overall computational overhead remains acceptable.
\begin{table}[htbp]
    \centering
    \caption{Fitting time under different target dimensionalities.}
    \label{tab:fitting_time}
    \begin{tabular}{ccccc}
        \toprule
        Target dimensionality & 256 & 512 & 1024 & 2048 \\
        \midrule
        Fitting time (s)      & 580 & 702 & 1476 & 2502 \\
        \bottomrule
    \end{tabular}
\end{table}

\section{Experiments on Small LLMs}
\subsection{Data Pool}
\textbf{Tulu3-SFT-Mixture \cite{DBLP:journals/corr/abs-2411-15124}.}
This open-source SFT dataset contains 939,344 high-quality instruction–response pairs, combining publicly available data with synthetic examples generated by GPT-4o in domains such as mathematics, coding, and precise instruction following. All samples undergo rigorous decontamination: any instance overlapping with established evaluation benchmarks is removed using 8-gram matching to ensure a fair and unbiased assessment.

\subsection{Experimental Settings and Results.} 
We also conduct additional experiments on smaller base models and datasets. In this setting, we select the top 40k most similar examples from the data pool as test samples. The base model is Qwen-2.5-7B, trained using LLaMA-Factory \cite{DBLP:journals/corr/abs-2403-13372}. All resulting models are subsequently evaluated using the original OpenCompass codebase. The training–evaluation pipeline is repeated three times, and we report the average performance across runs. For CRDS-R, unlike in the main setting, we use Qwen-2.5-7B as the encoder with the number of hidden states set to $H=4$. For the other baselines, all settings remain identical to those in our main experiments (Table \ref{main70k} and Table \ref{main100k}).

Preliminary results indicate that CRDS-based algorithms do not transfer effectively to smaller LLMs. Although CRDS-R demonstrates some advantage over RDS+, simpler and more intuitive methods—such as Length—emerge as surprisingly strong baselines in this setting.

\begin{table*}[htbp]
\centering
\caption{Experimental results on small models. See settings in Table \ref{exp_settings}.}
\label{small_model}
\setlength{\tabcolsep}{12pt}
\begin{tabular}{l lccccc}
\toprule
Base Model & Method & GSM & MMLU & MBPP & BBH & Average \\
\midrule
\multirow{6}{*}{Qwen2.5-7B}
 & Random      & 75.89 & 68.70 & 61.67 & 64.30 & 67.64 \\
 & Mid PPL     & 77.35 & 61.28 & 62.27 & 64.63 & 66.38 \\
 & Length      & 85.09 & 73.78 & 65.20 & 70.43 & 73.62 \\
 & RDS+        & 77.78 & 65.63 & 62.47 & 64.71 & 67.65 \\
 & CRDS-R & 78.95 & 72.78 & 62.67 & 64.52 & 69.73 \\
\bottomrule
\end{tabular}
\end{table*}

\subsection{The Effect of Average Pooling}
Average pooling is inherently a comparison method for CRDS-R. For the Qwen-2.5-7B base model and the Qwen-3-8B encoder, we show that average pooling does not exhibit the same level of sensitivity to layer depth as CRDS-R.
\begin{table}[htbp]
    \centering
    \caption{Effect of average pooling with different numbers of CRDS-R layers on GSM accuracy.}
    \label{tab:avg_pooling}
    \begin{tabular}{ccccccc}
        \toprule
        Number of layers in CRDS-R & 1 & 2 & 4 & 9 & 18 & 36 \\
        \hline
        Acc. on GSM (\%)           & 77.21 & 77.03 & 77.13 & 76.22 & 77.52 & 77.23 \\
        \bottomrule
    \end{tabular}
\end{table}

\section{Supplementary Settings of Experiments.}
We list all the supplementary settings for our experiments in Table~\ref{exp_settings}.
\begin{table}[htbp]
\centering
\small
\begin{tabular}{
  >{\centering\arraybackslash}p{0.25\linewidth}
  >{\centering\arraybackslash}p{0.7\linewidth}
}
\toprule
\textbf{Experiment} & \textbf{Settings} \\
\midrule
Table \ref{binarization} & The data pool is Ling-2.0-dev, with 50k samples selected.
The base model is Ling-mini-2.0, and the encoder is Llama-2-7B, using a truncation length of 2048 .
Each reported value represents the average of three parallel runs. \\
Figure \ref{whiten_ab} & The data pool is Ling-2.0-dev, with 50k samples selected.
The base model is Ling-mini-2.0, and the encoder is Llama-2-7B, using a truncation length of 4096.
Each reported value represents the average of five parallel runs. \\
Figure \ref{layer_ab} & The data pool is Ling-2.0-dev, with 50k samples selected.
The base model is Ling-mini-2.0.
Each reported value represents the average of five parallel runs. Llama-2-7B uses a truncation length of 2048, while the other three encoders use a truncation length of 4096. \\
Figure \ref{trunc_ab} & The data pool is Ling-2.0-dev, with 50k samples selected.
The base model is Ling-mini-2.0, and the encoder is Qwen-2.5-7B, with the number of hidden states $H=14$. Each reported value represents the average of five parallel runs. 
\\
Table \ref{raw_encoder}  &
The data pool is Ling-2.0-dev, with 50k samples selected.
The base model is Ling-mini-2.0, using a truncation length of 4096.
Each reported value represents the average of three parallel runs. \\
Figure \ref{fig:multilayer1} & The data pool is Ling-2.0-dev, with 50k samples selected.
The base model is Ling-mini-2.0, and the encoder is Qwen-2.5-7B, using a truncation length of 4096. Each reported value represents the average of three parallel runs.  \\
Figure \ref{fig:multilayer2} & The data pool is Ling-2.0-dev, with 50k samples selected.
The base model is Ling-mini-2.0, and the encoder is Llama-2-7B, using a truncation length of 4096, and was fine-tuned on 100,000 Ling-2.0-dev samples for 4 epochs. Each reported value represents the average of five parallel runs.\\
\bottomrule
\end{tabular}
\caption{Experimental settings used in our experiments.}
\label{exp_settings}
\end{table}

%%%%%%%%%%%%%%%%%%%%%%%%%%%%%%%%%%%%%%%%%%%%%%%%%%%%%%%%%%%%%%%%%%%%%%%%%%%%%%%
%%%%%%%%%%%%%%%%%%%%%%%%%%%%%%%%%%%%%%%%%%%%%%%%%%%%%%%%%%%%%%%%%%%%%%%%%%%%%%%

\end{document}